\RequirePackage[dvipsnames]{xcolor}     
\RequirePackage{tikz}					
\documentclass[iicol,pdflatex]{sn-jnl}  
\usepackage[noadjust]{cite}             
\usepackage[switch]{lineno}             

\jyear{2023}%
\usepackage[english]{babel}                     
\usepackage[utf8]{inputenc}                     
\usepackage[T1]{fontenc}                        
\usepackage{afterpage}                          
\usepackage{float}                              
\usepackage{siunitx}                            
\usepackage{bbding}                             
\usepackage[mathscr]{euscript}                  
\usepackage[b]{esvect}                          
\usepackage{soul}                               
\usepackage{enumitem}                           
\usepackage[nameinlink,capitalise]{cleveref}    
\usepackage[all]{hypcap}                        

\graphicspath{{Figures}}
\DeclareGraphicsExtensions{.pdf,.jpg,.jpeg,.png}

\NewCommandCopy\mycdot\cdot
\crefname{supfig}{Supplementary Fig.}{Supplementary Figs.}      
\Crefname{supfig}{Supplementary Figure}{Supplementary Figures}  

\newcommand{\restructure}[1]{\textcolor{RedViolet}{#1}}

\renewcommand{\vec}[1]{\protect\vv{\boldsymbol{\mathrm{#1}}}}
\newcommand{\vecsub}[2]{\protect\vv*{\boldsymbol{\mathrm{#1}}}{\,#2}}
\newcommand{\mat}[1]{\boldsymbol{\mathrm{#1}}}
\renewcommand{\B}{B}            
\newcommand{\I}{I}              
\newcommand{\cn}{\mathrm{c}}    
\newcommand{\sn}{\mathrm{s}}	

\let\diameter\varnothing

\usepackage{regexpatch}
\makeatletter
\xpatchcmd*{\@caption}{#3}{\textbf{#2}~#3}{}{}
\makeatother

\makeatletter
\newcommand*{\defercitelist}{}

\DeclareRobustCommand\defercite{%
  \@ifnextchar [{\@tempswatrue\@defercitex}{\@tempswafalse\@defercitex[]}}
\def\@defercitex[#1]#2{\leavevmode
  \let\@citea\@empty
  \@cite{\@for\@citeb:=#2\do
    {\@citea\def\@citea{,\penalty\@m\ }%
     \edef\@citeb{\expandafter\@firstofone\@citeb\@empty}%
     \listxadd{\defercitelist}{\@citeb}%
     \@ifundefined{b@\@citeb}{\hbox{\reset@font\bfseries ?}%
       \G@refundefinedtrue
       \@latex@warning
         {Citation `\@citeb' on page \thepage \space undefined}}%
       {\@cite@ofmt{\csname b@\@citeb\endcsname}}}}{#1}}

\newcommand*{\citeunsort@writecitation}[1]{%
  \if@filesw\immediate\write\@auxout{\string\citation{#1}}\fi}

\AtEndDocument{%
  \forlistloop{\citeunsort@writecitation}{\defercitelist}}
\makeatother

\makeatletter
\patchcmd{\ps@headings}
{\hbox to \textwidth{\hfill Springer Nature 2021 \LaTeX\ template\hfill}}
{\hbox to \textwidth{}}
{}
{}
\patchcmd{\ps@headings}
{\hbox to \textwidth{\hfill Springer Nature 2021 \LaTeX\ template\hfill}}
{\hbox to \textwidth{}}
{}
{}
\patchcmd{\ps@titlepage}
{\hbox to \textwidth{\hfill Springer Nature 2021 \LaTeX\ template\hfill}}
{\hbox to \textwidth{}}
{}
{}
\makeatother

\let\oldequation\equation
\let\oldendequation\endequation
\let\oldalign\align
\let\oldendalign\endalign

\renewenvironment{equation}
  {\linenomathNonumbers\oldequation}
  {\oldendequation\endlinenomath}
\renewenvironment{align}
  {\linenomathNonumbers\oldalign}
  {\oldendalign\endlinenomath}

\usepackage{xr}

\makeatletter
\newcommand*{\addFileDependency}[1]{
  \typeout{(#1)}
  \@addtofilelist{#1}
  \IfFileExists{#1}{}{\typeout{No file #1.}}
}
\makeatother

\newcommand*{\myexternaldocument}[1]{%
    \externaldocument{#1}%
    \addFileDependency{#1.tex}%
    \addFileDependency{#1.aux}%
}
\myexternaldocument{Supplementary}

\begin{document}

\title[Perching by Hugging]{\textbf{Crash-perching on vertical poles with a hugging-wing robot}}

\author*[1]{\fnm{Mohammad} \sur{Askari}}\email{mohammad.askari@epfl.ch}
\author[1]{\fnm{Michele} \sur{Benciolini}}
\author[1]{\fnm{Hoang-Vu} \sur{Phan}}
\author[1,2]{\fnm{William} \sur{Stewart}}
\author[3]{\fnm{Auke J.} \sur{Ijspeert}}
\author[1]{\fnm{Dario} \sur{Floreano}}

\affil*[1]{\orgdiv{Laboratory of Intelligent Systems}, \orgname{EPFL}, \orgaddress{\city{Lausanne}, \postcode{CH-1015}, \country{Switzerland}}}

\affil[2]{\orgdiv{Soft Flyers Group}, \orgname{Stony Brook University}, \orgaddress{\city{New York}, \postcode{11794}, \state{NY}, \country{USA}}}

\affil[3]{\orgdiv{Biorobotics Laboratory}, \orgname{EPFL}, \orgaddress{\city{Lausanne}, \postcode{CH-1015}, \country{Switzerland}}}

\abstract{
Perching with winged Unmanned Aerial Vehicles has often been solved by means of complex control or intricate appendages. Here, we present a simple yet novel method that relies on passive wing morphing for crash-landing on trees and other types of vertical poles. Inspired by the adaptability of animals' and bats' limbs in gripping and holding onto trees, we design dual-purpose wings that enable both aerial gliding and perching on poles. With an upturned nose design, the robot can passively reorient from horizontal flight to vertical upon a head-on crash with a pole, followed by hugging with its wings to perch. We characterize the performance of reorientation and perching in terms of impact speed and angle, pole material, and size. The robot robustly reorients at impact angles above \SI[detect-all=true]{15}{\degree} and speeds of \SIrange[detect-all=true]{3}{9}{\meter\per\second}, and can hold onto various pole types larger than \SI[detect-all=true]{28}{\percent} of its wingspan in diameter. We demonstrate crash-perching on tree trunks with an overall success rate of \SI[detect-all=true]{71}{\percent}. The method opens up new possibilities for the use of aerial robots in applications such as inspection, maintenance, and biodiversity conservation.}


\maketitle

\section*{Introduction}
\label{sec:introduction}

\vspace{0.3cm}
Winged Unmanned Aerial Vehicles (UAVs) are particularly suitable for long-distance missions, such as delivery, mapping, and search and rescue, as they offer higher endurance per mass compared to other types of UAVs \cite{vkumar2017energetics}. However, compared to winged flying animals, they have limited ability to land or perch on complex structures for tasks like inspection, manipulation, monitoring, or battery recharging \cite{mehanovic2017autonomous}. This limitation has spurred the development of control and mechanical systems to enable perching \cite{perching_review}. Inspired by avian perching \cite{graham2022avianperching}, most control-oriented studies have predominantly focused on pitch-up maneuvers and post-stall control for reducing speed at landing \cite{wickenheiser2008optimization,perched_landing_maneuver,paranjape2013novel}, and used either microspines to perch the winged UAV on a vertical wall \cite{microspine_wall_perching,mehanovic2017autonomous} or hooks to hang a glider on a cable \cite{powerline_perching,moore2014robust}. However, executing such maneuvers requires sensory systems embedded with control algorithms to ensure high level of accuracy within a short time frame. Furthermore, the rapid pitch-up maneuver is prone to potentially dangerous flight conditions due to reduced control effectiveness and aerodynamic stall at low speeds with high angles of attack.
\vspace{0.23cm}

In order to avoid the complex pitch-up maneuver, solutions with mechanical systems have been proposed as alternatives. Mechanical solutions for perching with hover-capable multicopters abound. Examples include passive avian-inspired claws \cite{roderick2021snag,quadrotor_avian_passive_claw_perching} and wrapping-arms \cite{quad_morphing_perching2023} for perching on branches, microspines for holding onto flat, rough, vertical surfaces \cite{scamp_robot}, modular landing gear \cite{quadrotor_perching_resting}, active \cite{luo2017vision} and passive \cite{tieu2016demonstrations, pounds2010hovering, thomas2016aggressive} compliant grippers, dry adhesive and fiber-based pads \cite{robobee_perching,jiang2014modeling, daler2013perching}, and spider-like perching using threaded anchors \cite{zhang2017spidermav}. However, there are fewer options available for winged UAVs. Anderson et al. \cite{anderson2009sticky} developed a simple adhesive-based mechanism that allows a fixed-wing UAV to attach to vertical surfaces upon head-on impact, followed by hanging from the anchor with a tether. Like most glue-based attachment concepts, this technique is surface-dependent and may not function effectively on damp or dusty surfaces. The study also does not provide any formal characterization of perching performance. Kova{\v{c}} et al. \cite{perching_needles_mechanism} proposed a system for perching a very lightweight microglider on walls, which consists of spring-loaded needles driven into the wall upon direct impact at speeds of up to \SI{4}{\meter\per\second}. However, this approach is more suitable for very lightweight robots with a weight of several tens of grams. For larger-scale systems, Stewart et al. \cite{stewart2021passive} introduced a passive perching claw that can mitigate kinetic energy to enable crash-perching at speeds up to \SI{7.4}{\meter\per\second}. Although the proposed claw design worked well for perching and hanging on small horizontal bars up to \SI{55}{\milli\meter}, it is not easily scalable or applicable to perching on vertical poles with larger diameters.

\begin{figure*}[ht]
\centering
\includegraphics[draft=false,width=\textwidth]{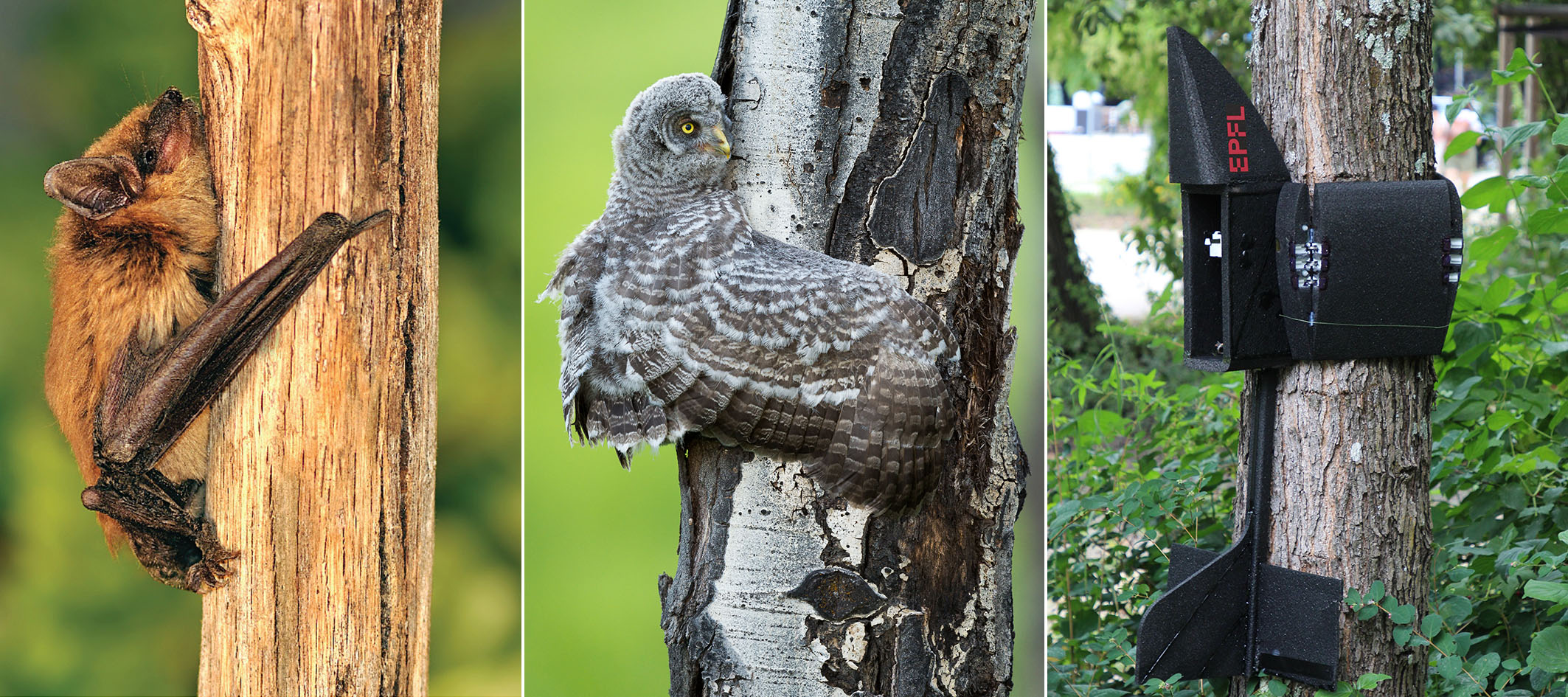}
\caption[Avian-inspired utilization of forelimbs for vertical perching on trees.]{A big brown bat (\emph{eptesicus fuscus}) holding onto a tree trunk using its wings and claws (left), a great grey owl (\emph{strix nebulosa}) fledging on its first day out of the nest wrapping its wings around a tree trunk to rest during climbing (center), and the PercHug robot perching vertically on a tree by hugging (right). Photo credits: \defercite{bat_photo,owl_photo}.}
\label{fig:animal vs robot}
\end{figure*}

Here, we propose a new method for the passive perching of winged UAVs on vertical poles, which are ubiquitous in man-made environments such as building scaffolding, electric towers, street lights, and utility poles. Perching on natural poles, such as trees, could also be helpful in biodiversity conservation or wildlife monitoring \cite{quad_morphing_perching2023,mintchev2023biosampling}. Geckos in their natural habitat exhibit a remarkable landing strategy on tree trunks. They crash head-first onto the trunk followed by a full body rotation, which is halted by the landing of their hind limbs and tail \cite{siddall2021tails}. Inspired by the gecko's touchdown technique, our proposed method incorporates an "upturned nose" element that allows passive reorientation from horizontal flight to a vertical attitude upon impact with the pole, thus foregoing control of complex pitch-up maneuvers at near-stall angles of attack. The UAV then leverages foldable, pre-loaded segmented wings, which are released through a latch system at impact, to wrap around vertical poles for perching. This behavior imitates that observed in certain flying animals (see \cref{fig:animal vs robot}) and eliminates the need for dedicated perching mechanisms that would increase body mass and complexity. We provide the design details of these elements, investigate the performance of inertial reorientation and wing wrapping induced by collisions, and validate the crash-perching capability on tree trunks using PercHug, a gliding-winged robot.

\section*{Results}
\label{sec:results}

\subsection*{Operating principle and robot design}
\label{sec:design}

\begin{figure*}[ht]
    \centering
    \includegraphics[draft=false,width=\textwidth]{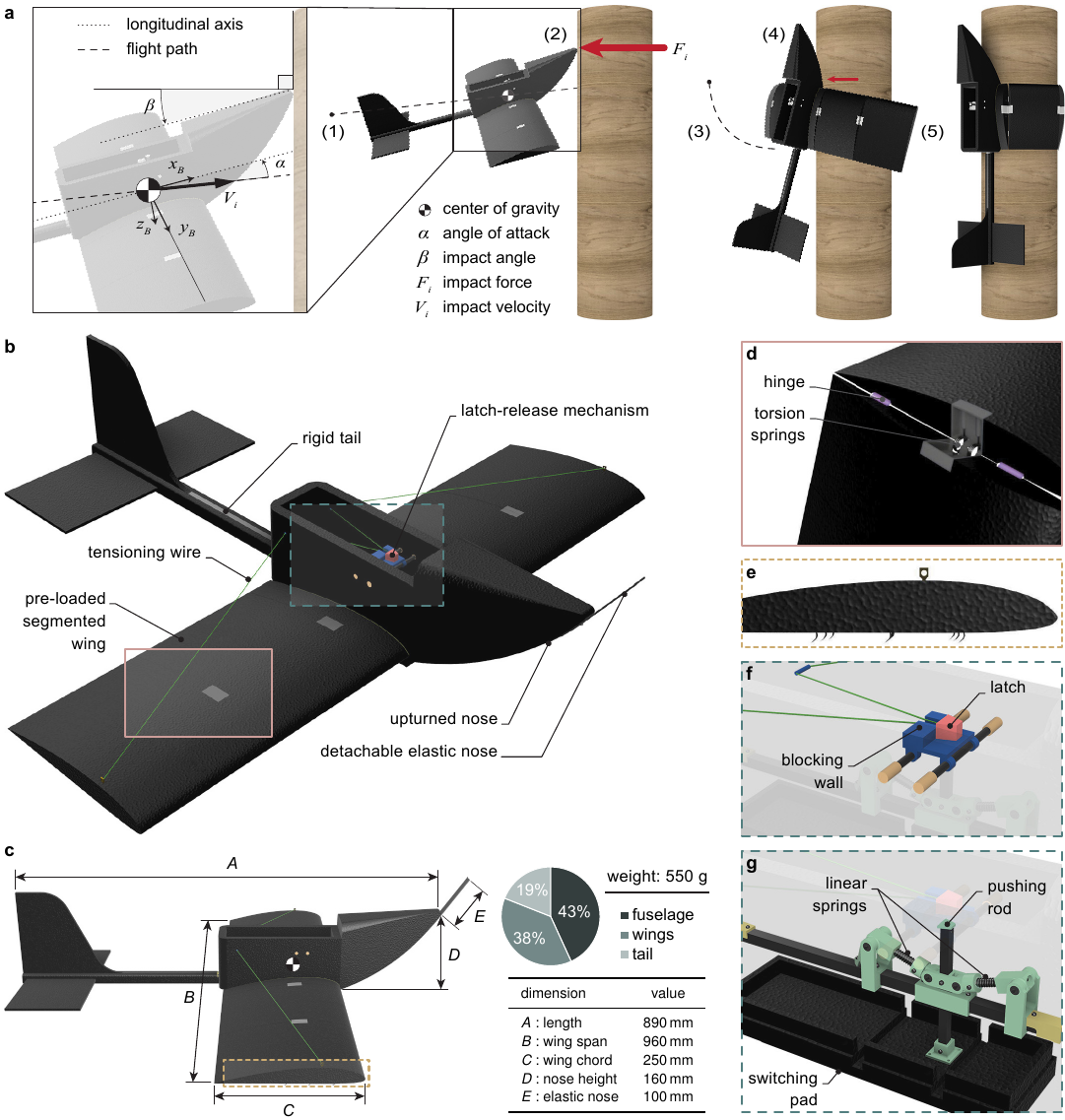}
    \caption[Perching strategy and architecture of the PercHug platform.]{\textbf{a} Operating principle of PercHug depicting the key steps of the perching maneuver: (1) gliding, (2) primary impact, (3) reorientation and wing release, (4) secondary impact, and (5) wing-wrapping. The red arrows represent the expected magnitudes of the impact forces, proportionally drawn. \textbf{b} Isometric view of PercHug showing different elements of the robotic platform. \textbf{c} Side view and physical properties of the robot. \textbf{d} Pre-loaded segmented wing interface in an open configuration. \textbf{e} Side view of the outermost wing segment highlighting the hooks. \textbf{f} Latching wing release mechanism (blue and red). \textbf{g} Backup bistable trigger (green).}
    \label{fig:design}
\end{figure*}

The complete perching maneuver occurs in a fraction of a second, within approximately \SI{200}{\milli\second} (see \cref{fig:design}a). It begins with the UAV flying directly toward a pole at certain speed and angle of attack to make a primary impact with the nose. The impact energy causes the robot to start rotating in the pitch direction and release its pre-loaded wings. The maneuver concludes with a secondary impact on the fuselage or tail to halt the rotation and the wings hugging the pole to hold the UAV in place. Only the correct sequence of events can lead to a successful landing on the pole.

The UAV design consists of the nose and wings design that can serve the dual purpose of flight and crash-perching (\cref{fig:design}b-g). This design strategy does not require additional hardware such as dedicated perching claws or feet to successfully perch the UAV. We present the principles behind the design of the integrated hardware of PercHug, encapsulated in an Expanded PolyPropylene (EPP) foam body, with a ready-to-perch weight of \SI{550}{\gram} and a wingspan of \SI{96}{\centi\meter} (\cref{fig:design}; see \hyperref[sec:fabrication]{Methods} for the fabrication details). 

\begin{figure*}[ht]
    \centering
    \includegraphics[draft=false,width=\textwidth]{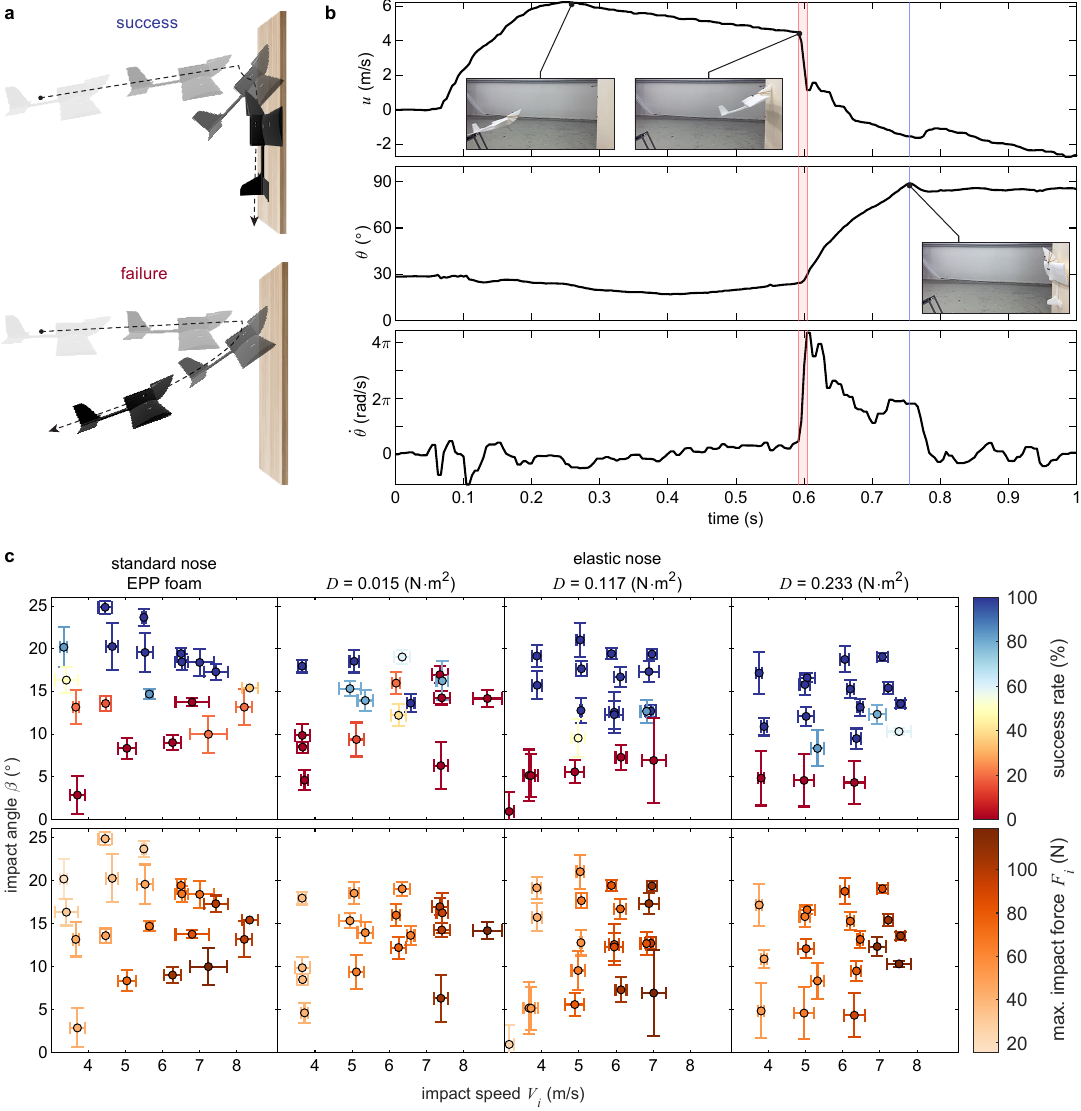}
    \caption[Reorientation maneuver and performance with four different noses.]{\textbf{a} Illustration of the concepts of unsuccessful reorientation, where the UAV bounces off the wall after impact, and a successful one in which it reaches a vertical orientation while making secondary contact with the wall. \textbf{b} Time evolution of the UAV's translational velocity $u$, pitch angle $\theta$, and pitch rate $\dot{\theta}$ for a sample trial (see \cref*{fig:kinematics} and {\hypersetup{hidelinks}\hyperref[sec:flight kinematics]{Methods}} for the definitions of the state variables). The red region shows the duration of the primary impact, and the blue line corresponds to the time of maximum pitch for a successful reorientation. \textbf{c} Characterization results of the UAV reorienting from horizontal to vertical configuration. The plots show variations in success rate and mean primary impact force with impact angle and speed for four different types of noses.}
    \label{fig:reorientation}
    \vspace{-0.1cm}
\end{figure*}

\begin{figure*}[ht]
    \centering
    \includegraphics[draft=false,width=\textwidth]{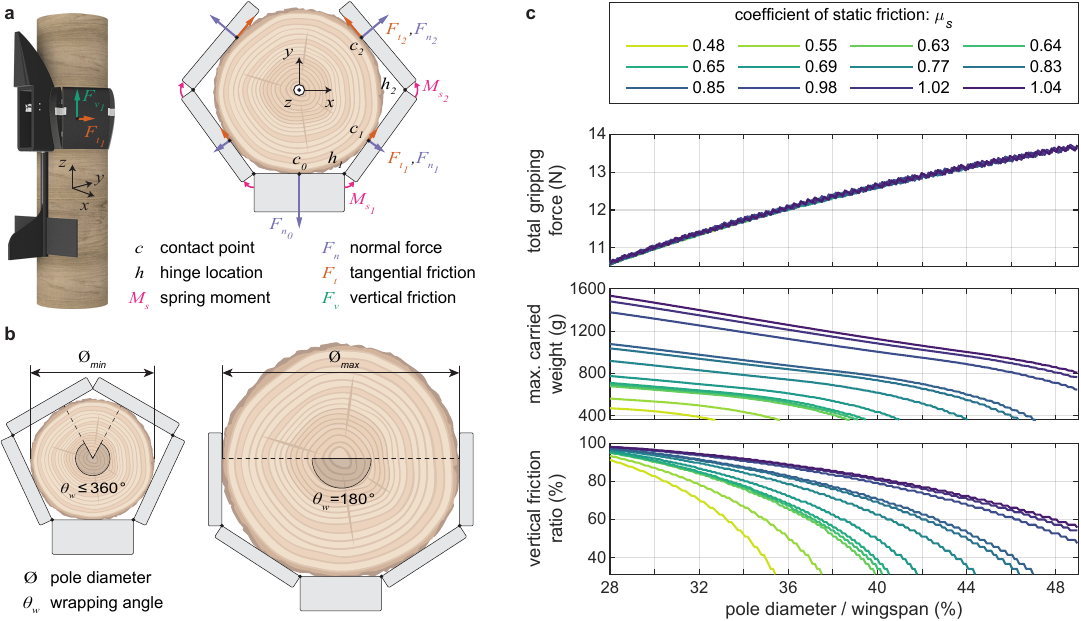}
    \caption[Static wing-wrapping model and pole gripping performance.]{\textbf{a} Free-body diagram used for static modeling of the robot perched on a pole, shown in isometric and top views (refer to {\hypersetup{hidelinks}\hyperref[sec:model]{Methods}} for further details). \textbf{b} The theoretical minimum and maximum pole diameters the robot can perch on. \textbf{c} Simulation results of the static model showing variations in net squeezing force by the wings, maximum static payload capacity, and friction split with pole size and material. The static friction coefficients correspond to the poles used for the actual static experiments, while the diameter range is defined by the minimum and maximum values.}
    \label{fig:static model}
    \vspace{-0.2cm}
\end{figure*}

The novel design of the upturned nose shape (\cref{fig:design}b) enables the robot to passively reorient from horizontal flight to the vertical configuration needed for perching. The primary impact force at the tip of the nose, which is large in magnitude, generates a moment about the center of gravity (COG) to yield rotation, implying that the relative placement of the nose tip with respect to the COG plays a crucial role in reorientation performance (\cref{fig:design}c). In addition to the upturned nose shape, we also studied the reorientation performance by extending the tip of the nose with a flexible flat carbon bar (\cref{fig:design}b and c). 

The vehicle is equipped with foldable wings that have three hinged segments. One segment is attached to the fuselage, while the other two can bend in the ventral direction to wrap around the pole (\cref{fig:design}d). Two torsion springs with a stiffness of \SI{1.36}{\newton\milli\meter\per\degree} are placed at the interface between the two segments and are pre-loaded during flight. Upon impact, the springs are released and cause the segments to fold and press against the pole. This gripping force, combined with the friction along the vertical axis resulting from the gravitational force, keeps the robot attached to the pole.

The outermost segments of the wings can be equipped with nine removable hooks (\cref{fig:design}e) to help engage with rough surfaces (such as the bark of a tree). A tensioning wire keeps the wings open and straight during flight. The wire connects the two tips of the wings to a latch in the fuselage (\cref{fig:design}b), and its length is adjusted to provide a dihedral angle of about \SI{5}{\degree} in order to improve the lateral stability of the aircraft.

The latching mechanism (\cref{fig:design}f) holds the wire in tension during flight and passively releases it upon impact. It comprises a fixed piece (shown in blue) firmly attached to the fuselage and a latch (shown in red) through which the wire passes. The pulling force exerted by the wing springs on the wire, connected to the latch, is blocked by the vertical wall of the fixed blue piece during flight. Wings are released when the latch pops over the blocking wall. We can adapt the release time by adjusting the wall height. With a height of \SI{5}{\milli\meter}, the latch is released upon primary impact with the pole (\cref{fig:design}a-2) due to the shock taken by the airframe. If the walls are \SI{10}{\milli\meter} high, the wings are not released at primary impact but are unlatched at the secondary impact once fully reoriented vertically (\cref{fig:design}a-4). This is made possible with the backup bistable trigger (\cref{fig:design}g), which operates similarly to the mechanism in \cite{perchflie_bistable}. It sits right beneath the latch and connects to a switching pad that extends out of the fuselage during flight, aided by a pair of compressed springs. A secondary impact with the underside of the fuselage triggers the release passively. The impact pushes the pad inward, causing the bistable mechanism to switch its position. As a result, the pushing rod moves upward and releases the latch by popping it over the blocking wall, freeing the wings.

\subsection*{Inertial reorientation}
\label{sec:reorientation}

We decoupled the reorientation and pole-hugging problems to independently study nose selection and wing design. Here, we primarily investigate how the upturned nose, with and without elastic nose extensions (see \cref{fig:design}), performs in reorienting the robot at impact. The rationale behind using flexible noses is to examine whether they enhance reorientation by extending the impact moment arm with respect to the COG, decreasing the maximum impact force taken by the airframe, and redirecting the force vector more effectively toward sliding up and attaching to the surface.

We consider a reorientation successful if the UAV reaches a vertical orientation and makes secondary contact with the wall (\cref{fig:reorientation}a). In contrast, a failure means a rebound off the surface after the primary impact and a lack of secondary impact. The tracking data for a successful reorientation case is illustrated in \cref{fig:reorientation}b. The primary impact phase (highlighted in red) indicates the start of the reorientation and is identified by the sharp drop in translational speed and the sharp increase in pitch rate. The end of the reorientation maneuver is also considered to be the point of maximum pitch angle (marked by the blue line). The impact speed $V_i$ and relative impact angle $\beta$ (shown in \cref{fig:design}a) are estimated from the moment of primary impact. Notably, $\beta$ corresponds to the robot's pitch at impact, given that the wall is placed vertically.

The reorientation success rate depends primarily on the impact angle rather than on variations in impact speed. This behavior is clearly seen in \cref{fig:reorientation}c, where reorientations fail below a certain impact angle threshold, regardless of the type of nose used. With the standard upturned nose, the vehicle successfully reorients for impact angles above about \SI{15}{\degree} at speeds of \SIrange{3}{9}{\meter\per\second}. Such speeds match cruising speeds of comparable vehicles in size and weight, such as \cite{perched_landing_maneuver}. In comparison, reorientation performance improves with increased flexural rigidity $D$ of the elastic noses (see \hyperref[sec:flexural rigidity]{Methods} for the definition of flexural rigidity). This trend is seen by the successful reorientations at lower impact angles, reaching up to a minimum of \SI{8}{\degree} for the stiffest nose. The results thus indicate that the elastic nose extension with $D=\SI{0.233}{\newton\meter\squared}$ is the best-performing one out of the nose types tested. Despite the differences among the success rates, the duration of the reorientation maneuver remains consistent for different speeds, angles, and nose types. We define this duration as the time from primary impact to \SI{90}{\degree} pitch and measure it as an average of $196\pm$\SI{59}{\milli\second}.

Unlike the success rate, the primary impact force $F_i$ (shown in \cref{fig:design}a), which is estimated from the speed profile, is linearly proportional to the impact speed $V_i$ (see \hyperref[sec:flight kinematics]{Methods} for mathematical formulation). Its peak value varies between about \SI{20}{\newton} to \SI{120}{\newton} for speeds of \SIrange{3}{9}{\meter\per\second} (\cref{fig:reorientation}c). In the case of the standard upturned nose, the primary impact force is also a function of the impact angle, i.e., the impact force increases as the impact angle decreases. However, the correlation is unclear for the other nose types possibly due to  effect of the nose flexibility. Although the elastic nose with $D=\SI{0.233}{\newton\meter\squared}$ improves the success rate at lower impact angles compared to the standard upturned nose, \cref{fig:reorientation}c also shows that they share similar amounts of impact force over the range of tested impact speeds. The improved success rate may be a result of a longer moment arm from the impact point to the COG of the robot (\cref{fig:design}a). It is also expected that there will be a linear correlation between weight and impact force (\cref{eq:impact force}). The data presented here can help estimate the impact force for robots of similar sizes at different weights, which is valuable for the mechanical design and structural analysis of the airframe.

\subsection*{Static perching}
\label{sec:static perching}

\begin{figure*}[ht]
    \centering
    \includegraphics[draft=false,width=\textwidth]{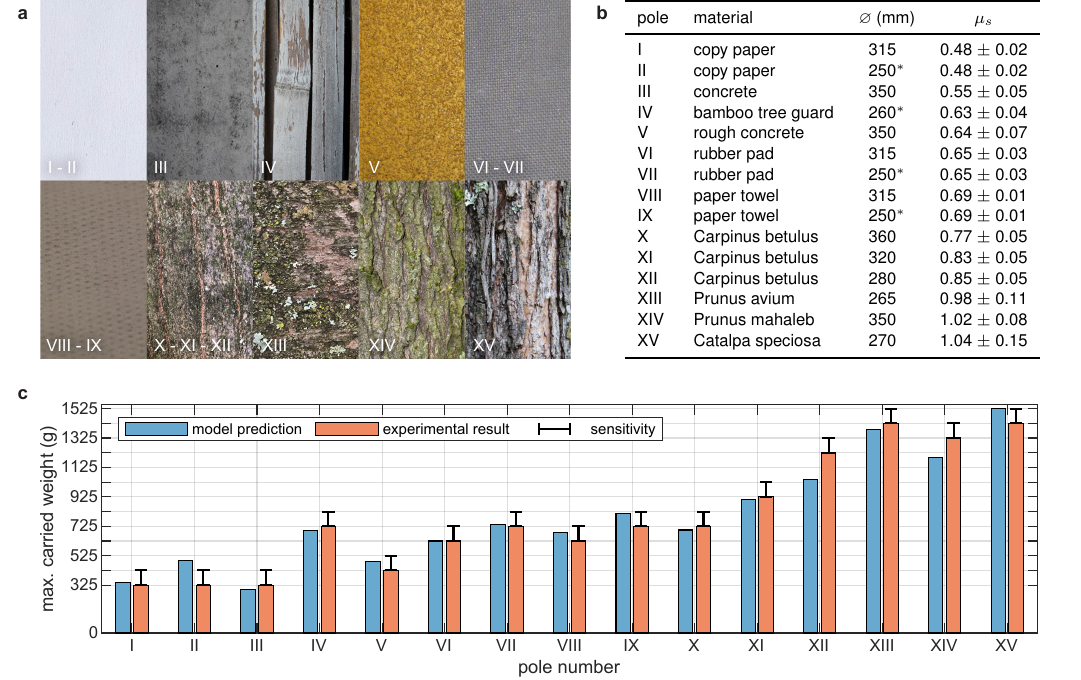}
    \caption[Model validation with static perching experiments.]{\textbf{a} Close-up pictures of the surfaces of the poles used in the static perching experiments. \textbf{b} List of poles and their specifications in order of increasing friction coefficient. The "*" symbol denotes poles with a diameter smaller than the model's predicted minimum value of \SI[detect-all]{265}{\milli\meter}. These cases were analyzed since the model was found to be valid even outside the previously mentioned diameter range, provided that the considered diameter is close to the limit. \textbf{c} Model prediction and results of the real-world static experiments. The measured values start at the weight of the prototype alone (\SI[detect-all]{325}{\gram}) and increase in increments of \SI[detect-all]{100}{\gram} (see {\hypersetup{hidelinks}\hyperref[sec:static perching experiments]{Methods}} for more details). The insignificant discrepancies between experiment and simulation results in cases II, XII, and XIV can be attributed to minor errors due to the non-uniformity of tree barks (XII and XIV) and possible discrepancy of the stiffness coefficient of the torsion springs for small poles (case II) that exceed the manufacturer values for linear operation.}
    \label{fig:static perching}
\end{figure*}

\begin{figure*}[ht]
    \centering
    \includegraphics[draft=false,width=\textwidth]{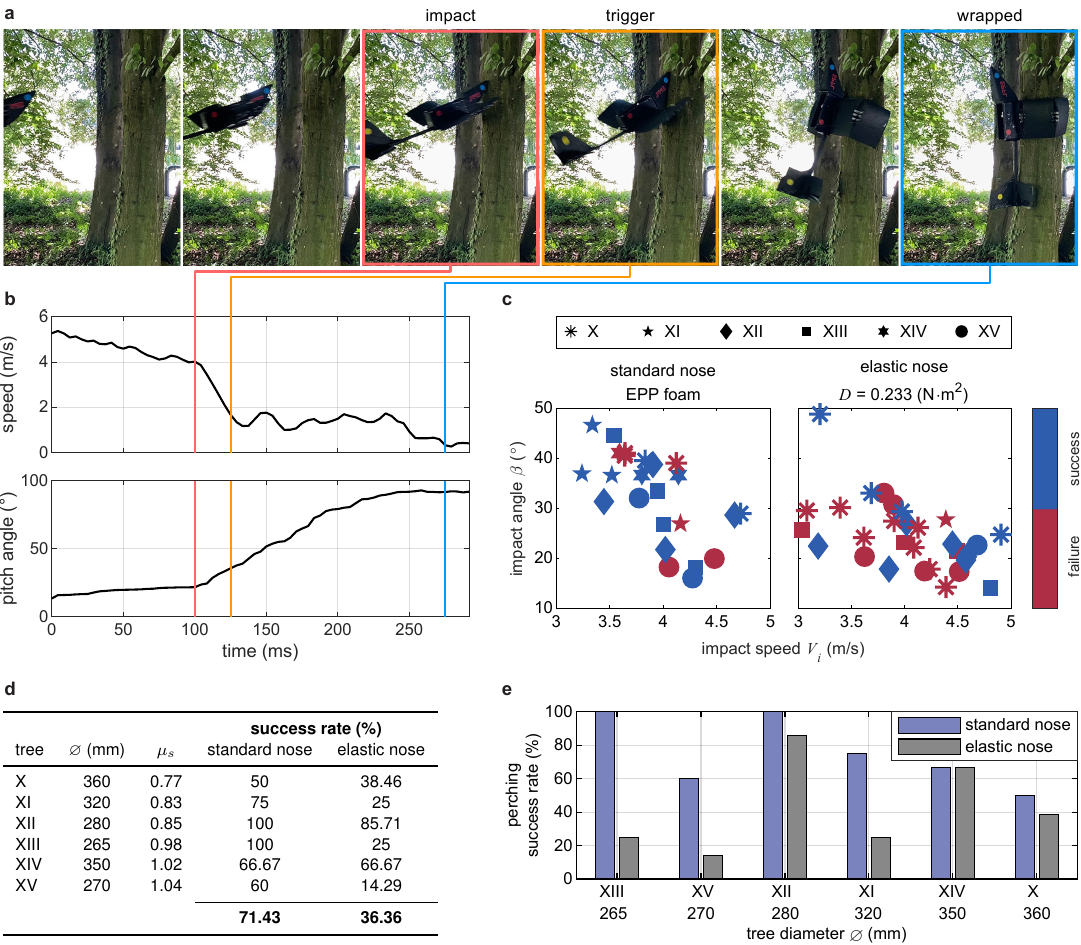}
    \caption[Dynamic perching performance on trees with two different nose configurations.]{\textbf{a} Snapshots of a crash-perching experiment with PercHug on a tree trunk (XII), captured from high-speed footage (Supplementary Video). \textbf{b} The speed and pitch angle variation during the dynamic perching maneuver (see {\hypersetup{hidelinks}\hyperref[sec:dynamic perching experiments]{Methods}} for further information). PercHug glides and settles at a speed of around \SI[detect-all]{4}{\meter\per\second} with a pitch of \SI[detect-all]{20}{\degree} before impact with the tree. The sharp decrease in speed and increase in the pitch angle indicate the beginning of the reorientation phase. Triggering occurs right after the primary impact and the tensioning wire is released. The wings are seen wrapping and closing around the tree during reorientation. The wrapping phase concludes when the wingtips make contact with the rear surface of the tree trunk. \textbf{c} Success map of dynamic perching on various trees with the standard upturned nose and extended elastic nose. \textbf{d} List of trees, their specifications, and the corresponding success rate of perching on each one. \textbf{e} The plot shows the success rate with increasing tree diameter.}
    \label{fig:dynamic perching}
    \vspace{-0.1cm}
\end{figure*}

The sizing and segmentation of the folding wings define the range of pole diameters on which the robot can perch. In order to understand the most suitable dimensions, we developed a wing-wrapping model (\cref{fig:static model}a; see \cref*{fig:flowchart} and \hyperref[sec:model]{Methods} for details), and validated it by conducting pole-hugging experiments with the predicted wing design. The model is applicable to static perching, which refers to the situation when the wings are already wrapped around the vertical pole.

In contrast to insects, which use dry or wet adhesion or claws to hold onto tree surfaces, larger animals with articulated limbs leverage interlocking methods by encircling over half the trunk with their forelimbs (see \cref{fig:animal vs robot}) \cite{tree_climbing_in_nature}. Similarly, here we assume that the maximum pole diameter on which the UAV can perch corresponds to a wing-wrapping angle ($\theta_w$) of \SI{180}{\degree}, and the minimum pole diameter corresponds to a size that prevents the two wingtips from overlapping (\cref{fig:static model}b). For the wings to remain wrapped, the normal and tangential reaction forces of the outermost segments ($F_{n_2}$, $F_{t_2}$), which have components directed in $+y$, must be sufficiently large to counteract the $-y$ directed forces on the fuselage and other segments. In the event that a pole of larger diameter is employed ($\theta_w < \SI{180}{\degree}$), it follows that almost no force exerts in the $+y$ direction, which inevitably causes the robot to slide off the pole.

The model takes into account various body and wing characteristics, including wingspan, fuselage width, the number and sizes of the folding-wing segments, and the stiffness of torsion springs, along with the coefficient of static friction between EPP and the pole material. These factors help estimate the range of diameters on which the UAV can perch and the maximum static payload. The range of pole diameters suitable for perching mainly scales with the wingspan and spans from approximately \SIrange{28}{50}{\percent} of it, as predicted by the model. For the selected \SI{960}{\milli\meter} wingspan with two folding segments per wing, this corresponds to diameters of \SIrange{265}{470}{\milli\meter}.

The model was also used to study variation in gripping force and maximum static payload with the pole diameter $\diameter$ and coefficient of static friction $\mu_s$. These two factors critically influence the wing-wrapping performance. The analysis revealed a positive correlation between the net gripping force and the pole diameter (\cref{fig:static model}c). This can be attributed to the greater compression of the torsion springs between the wing segments, leading to a stronger squeezing force on a wider pole. However, the maximum weight the UAV can support before sliding down the pole reduces as the diameter increases. This is due to the sharp decrease in the fraction of the total friction force acting along the vertical axis, which outweighs the increase in pressing force on wider diameters. As a result, although the total friction force increases with an increase in pole diameter, its vertical component ($F_v$) decreases, limiting the amount of mass the UAV can hold. Concerning the effect of coefficient of static friction, an increase in $\mu_s$ leads to an improvement in overall static perching performance, as expected.

We utilized the wing-wrapping model to determine suitable dimensions for the segments, and validated the selected wing design with experiments of a physical prototype attached to poles of different sizes and materials. The surface texture and specifications of the tested poles are listed in \cref{fig:static perching}a and b. Our symmetrical wing segmentation approach yielded folding segments of size \SI{195}{\milli\meter} (see \cref*{fig:configurations} and \hyperref[sec:segmentation]{Methods} for the segmentation study). The results exhibit a strong agreement between the measured values and the model predictions (\cref{fig:static perching}c). As predicted by the model, the experiments show that the maximum payload increases as the diameter decreases, assuming the surface material remained the same (evidenced by pairs I-II, VI-VII, and VIII-IX). Despite the variation in diameter, an overall upward trend in payload is observed from left to right, highlighting the positive effect of increasing coefficient of friction on payload capacity.

\subsection*{Experimental validation with PercHug}
\label{sec:dynamic perching}

We leveraged the insights gained from the reorientation studies and the static perching model to size and characterize the hug-perching performance of PercHug when hand-launched against trees. PercHug weighed \SI{550}{\gram}, including the unlatching mechanism, bistable backup trigger, and hooks, as well as a reinforced tail and body (\cref{fig:design}b-g) to enhance durability during multiple crash-perching tests. In these experiments, we tested PercHug equipped with and without the extended elastic nose of $D$\,=\,\SI{0.233}{\newton\meter\squared} (corresponding to the best reorientation performance) on the six trees used in the static perching experiments (trees X-XV in \cref{fig:static perching}).

In all dynamic perching experiments, we hand-launched the robot toward the trees. A perching trial was considered successful if it involved four distinct phases of gliding, reorienting at impact, wrapping the wings, and staying perched on the tree (see \cref{fig:dynamic perching}a and Supplementary Video). We first investigated the effect of unlatching time on perching by adapting the latch mechanism to release at either the primary or secondary impact using the bistable trigger (\cref{fig:design}a, f, and g). Preliminary studies revealed that the secondary impact release strategy resulted in nearly always unsuccessful trials, regardless of the nose type. This was due to the robot falling off the tree shortly after the secondary impact, rather than just falling straight down, and as a result, it was pushed away from the tree making it almost impossible to wrap around. Conversely, when using the primary impact release strategy, the robot displayed more robust perching capabilities.

Analyzing the video and tracking data from successful perching trials (see \hyperref[sec:dynamic perching experiments]{Methods} for further information) reveals that the release mechanism was triggered very shortly after the primary impact, with about a \SI{20}{\milli\second} delay (\cref{fig:dynamic perching}a and b). Additionally, the wrapping phase concluded almost simultaneously with the end of the reorientation phase, which is characterized by the instance of reaching an approximately \SI{90}{\degree} pitch. In this specific test, the durations of reorientation and wrapping, measured from the moment of impact, were approximately \SI{150}{\milli\second} and \SI{180}{\milli\second}, respectively. These results confirm the rapid dynamics of the perching maneuver and underscore the importance of timing the unlatching strategy to release at the primary impact. Furthermore, The rigid tail of PercHug plays a crucial role in halting the reorientation maneuver upon contact with the trunk when reaching a vertical position. This is evident from the pitch data, where the pitch increase ceases and levels off at around \SI{90}{\degree} due to the tail's contact (\cref{fig:dynamic perching}b). In trials where the robot did not make contact with its tail, it continued to reorient beyond \SI{90}{\degree}, resulting in unsuccessful landing maneuvers. This finding aligns with the research conducted by Siddal et al. \cite{siddall2021tails} on the impact of tails in tree-perching geckos, which emphasized the advantages of longer-tailed robots when crash-landing at speeds between \SIrange{3}{5}{\meter\per\second}.

PercHug successfully demonstrated crash-perching capability on all trees for impact speeds $V_i$ ranging from \SIrange{3}{5}{\meter\per\second} and relative impact angles $\beta$ above \SI{15}{\degree}, regardless of the nose type (\cref{fig:dynamic perching}c). The success maps illustrate the distribution of successful and failed perching trials across different trees. These experiments occurred to have very similar average impact speeds of $4.1 \pm \SI{0.7}{\meter\per\second}$ for both nose configurations.

Perching was more successful with the standard upturned nose compared to the elastic nose on every individual tree (\cref{fig:dynamic perching}d). The standard nose configuration yielded an overall perching success rate of \SI{71}{\percent} across all trials, almost double that of the elastic nose (\SI{36}{\percent}). While the elastic nose improved reorientation by facilitating the nose slide-up on the pole (\cref{fig:reorientation}c), it hindered surface attachment by creating a larger gap between the robot and the tree, making it harder for PercHug to wrap around. In other words, some of the kinetic energy was stored as elastic energy inside the bent nose and released, effectively pushing the robot away from the tree. The standard nose configuration exhibited a \SI{30}{\percent} faster triggering time compared to the experiments with the elastic nose, which potentially contributed to its higher success rate. The average trigger delays from impact were $26 \pm \SI{6}{\milli\second}$ and $37 \pm \SI{15}{\milli\second}$, for the standard and elastic noses, respectively. In successful trials, the respective mean wrapping times were $143 \pm \SI{44}{\milli\second}$ and $155 \pm \SI{30}{\milli\second}$, with a variation of less than \SI{8}{\percent} for the standard and elastic noses. The comparable wrapping times imply that this parameter is likely a characteristic of the wings and unaffected by other UAV components.

With the exception of tree XV, experimental results with the standard nose also indicate that wider trees lead to lower success rates (\cref{fig:dynamic perching}e). This behavior aligns well with the trend predicted by the model (\cref{fig:static model}c) and observed in static perching experiments. While the coefficient of static friction $\mu_s$ and pole diameter $\diameter$ are equally crucial factors for static perching, the diameter of the tree plays a more significant role in dynamic perching success, provided that the friction is sufficient for the robot to maintain its grip.

\section*{Discussion}
\label{sec:discussion}

We have presented a novel bio-inspired method to passively crash-land on vertical poles and trees with winged robots. While flying animals substantially reduce their kinetic energy at landing by wing flapping, specialized gliders, like flying squirrels and geckos, land on trees at high speeds and endure significant forces utilizing their limbs or head \cite{roderick2017touchdown}. We took inspiration from flying geckos, which exhibit head-first crash-landing at speeds between \SI{5}{\meter\per\second} and \SI{7}{\meter\per\second} \cite{siddall2021tails}. In a similar manner, our proposed upturned nose robustly transforms the impact kinetic energy to a passive reorientation maneuver at \SIrange{3}{9}{\meter\per\second}, thus eliminating the need for a complex pitch-up maneuver to bleed off speed before perching. Additionally, geckos have been observed to reach the target with their body pitched upward between \SI{8}{\degree} and \SI{24}{\degree} \cite{siddall2021tails}. Our experimental results align with these findings, showing that a minimum impact angle of \SI{15}{\degree} with the upturned nose (or \SI{8}{\degree} with an extended elastic nose) is crucial for successful reorientation. Although these angles are small, optimizing the position of the nose tip relative to the COG can enable flight and reorientation at smaller angles of attack. The impact force, speed, and angle characterization results, coupled with the robot's weight and inertia data, provide the foundation for developing a dynamic model to further explore design possibilities at different speeds, angles of attack, and pole dimensions.

Large size mammals lacking sharp claws, and less commonly some flying animals (\cref{fig:animal vs robot}), use the interlock method when climbing or resting on tree trunks by encircling at least half of the trunk with their forelimbs \cite{tree_climbing_in_nature}. This behavior inspired the design of our foldable, pre-loaded segmented wings that serve the dual purposes of flight and perching by securely wrapping around poles after impact. Our static perching model and tests similarly showed that we are constrained by a wrapping angle of \SI{180}{\degree}, and that the range of pole sizes the robot can perch on is solely determined by the choice of wingspan and segmentation. Moreover, the pole diameter and the coefficient of static friction are discovered to be equally important factors for maintaining a secure perch. The results highlighted that the static payload capacity varies inversely with the diameter and directly with friction. These findings are in line with those observed in nature reporting the interlock fastening method's dependency on the angle of the frictional force, which leads to a reduced gripping force when animals climb larger branches \cite{tree_climbing_in_nature, primates_climbing_friction}. We also investigated the sizing relationship and segmentation of the wings through our model, providing insights that can aid in the design of improved wings and the prediction of performance for robots of different dimensions.

PercHug's experimental results have brought us significant insights into dynamic perching on trees. Two crucial factors have been identified as determinants of perching success. First, the rapid dynamics of the perching maneuver, occurring in under \SI{200}{\milli\second}, underscore the importance of precise timing for wing release. Optimal results are achieved by releasing the wings at the beginning of the reorientation phase upon primary impact, while releasing them at the end of reorientation by the secondary impact leads to almost complete perching failure. 
Second, tree diameter has emerged as a more dominant factor than the friction coefficient in determining the success rate. Consistent with the predictions of the static model, perching improves on smaller tree sizes within the acceptable range of diameters compared to larger ones. It is also worth noting that despite the elastic nose's potential for improved reorientation, its detrimental push-away effect resulted in a lower perching success rate (\SI{36}{\percent}) compared to the standard upturned nose (\SI{71}{\percent}). Future developments based on this work will integrate avionics to enable autonomous flight and perching. The incorporation of a propulsion system holds potential for further enhancing perching success by assisting during the landing maneuver. Ongoing efforts involve sensor-based pole detection and wing release, an active grip loosening control for propeller-assisted climbing, and unperching to complete a full mission cycle.

We firmly believe that our study lays a foundation for advancing perching technologies and paves the way for the development of highly versatile robotic systems tailored to diverse applications. Such robots could be deployed for inspection tasks in complex industrial environments or tall buildings, enabling close-up examinations without the need for scaffolding or risky human interventions. In the field of infrastructure inspection and maintenance, perching would enable them to access challenging locations with ease, like electric or cellular towers, allowing for the inspection of power grid systems and communication framework by assessing equipment functionality. Perching on lamp posts or street signs in urban environments could enhance surveillance and security systems. In environmental monitoring applications, these versatile robots could perch on trees to gather data on biodiversity, habitat conditions, and ecological changes. Additionally, they could serve as valuable tools for studying wildlife behavior, enabling non-intrusive data collection to support wildlife conservation efforts. The possibilities are vast, and as perching solutions evolve, these robotic systems will find their place in numerous domains, reshaping the way we interact with and benefit from intelligent aerial machines.

\section*{Methods}
\label{sec:methods}

\subsection*{Robot fabrication}
\label{sec:fabrication}

The PercHug robot (shown in Fig. \cref{fig:design}) is made of three main material groups: Expanded Polypropylene (EPP) flexible lightweight foam, 3D-printed Tough Polylactic Acid (PLA), and fiber reinforced carbon bars. The upturned nose, wings, tail, and the soft shells of the fuselage are made out of multiple patterns cut out of EPP using a hot wire foam cutting tool. Each element is created separately and then assembled at the final stage.

The fuselage shells are glued together with UHU por and encapsulate the internal skeleton made out of an \SI{8}{\milli\meter} x \SI{8}{\milli\meter} pultruded carbon fiber square tube. The pieces of the latching wing release mechanism, which sits over two \SI{4}{\milli\meter} round carbon tubes at the top of the fuselage, as well as the backup bistable trigger are 3D-printed out of Tough PLA. The bistable trigger's pushing rod (\SI{4}{\milli\meter} x \SI{4}{\milli\meter} square carbon bar) connects to the bottom shell of the fuselage (switching pad). The rest of the system assemble through connecting the pieces along with two linear springs and eight bearings, whose interfaces are directly glued to the fuselage's carbon bar.

Three identical touph PLA 3D-printed pieces hold M2 threaded brass inserts, which are inserted and glued with five minute Epoxy to the underneath of the foam nose for the attachment of the elastic extensions. The foam nose is then directly glued with UHU por to the front shell of the fuselage. The tail is reinforced with a \SI{6}{\milli\meter} x \SI{6}{\milli\meter} pultruded carbon fiber square tube, to which its foam pieces are glued with five minute Epoxy. The square tube is tightly fitted into the main fuselage carbon bar, making the tail replaceable in case of damage. Each wing is cut into corresponding segment sizes, and every two segments are held together with two \SI{20}{\milli\meter} x \SI{36}{\milli\meter} plastic hinges, glued to their flat bottom side. A square channel is cut at every segment's interface, where the housing for the two torsion springs is fitted and glued. We use five minute Epoxy to interconnect the wings parts. Additionally, nine fishing hooks are hot-glued to the bottom side of each outermost segment. The left and right wings are brought together and their fixed segments are assembled to the fuselage using two cross round carbon tubes with a diameter of \SI{5}{\milli\meter}. Lastly, two eye screws are fixed to 3D printed pieces on the wing tips through which we pass the Dyneema (0.2 mm, 20 kg) that connects to the latch in the fuselage.

\subsection*{Reorientation experiments}
\label{sec:reorientation experiments}

The experimental setup shown in \cref*{fig:reorientation setup} was built to characterize the reorientation performance based on impact speed $V_i$, impact angle $\beta$, and nose type (see \cref{fig:design}). We tested four different noses, one with the standard upturned foam shape and three with extended elastic beams of different cantilever flexural rigidity $D$ (see \nameref{sec:flexural rigidity}), to test the effectiveness of the flexible noses. We used a \SI{220}{\gram} fixed-wing glider of dimensions given in \cref{fig:design}c, equipped solely with these nose types to eliminate potential effects due to other UAV elements, and launched the robot with a bungee-powered catapult with adjustable speed and angle toward a vertical wall. For each nose, we tested the vehicle at speeds of approximately \SIrange{3}{9}{\meter\per\second} and positive pitch angles of up to \SI{25}{\degree}. We conducted a total of more than $100$ experiments, each repeated with at least five trials, tracked the robot's position and attitude in space by an OptiTrack motion capture system (see \nameref{sec:flight kinematics} for the derivation of other states), and recorded slow-motion videos with a camera at \SI{240}{\hertz} (\cref*{fig:reorientation setup}).

\subsection*{Elastic noses}
\label{sec:flexural rigidity}

Flexural rigidity or bending stiffness of a cantilever beam, which is fixed at one end and free at the other, is the measure of its resistance to bending subject to an external load. It is measured as the product of the modulus of elasticity $E$ and area moment of inertia $I$ of the beam. For the elastic noses, we used flat pultruded carbon fiber profiles with rectangular cross-sections, whose flexural rigidity are calculated by
\begin{equation}
D = EI = \dfrac{E b h^3}{12} \ , \label{eq:flexural rigidity}
\end{equation}
where $b$ and $h$ are the width and height of the cross-section, respectively. A higher flexural rigidity indicates a stiffer beam, which can withstand larger bending loads and experience smaller deflections.

\subsection*{Flight kinematics and impact force estimation}
\label{sec:flight kinematics}

Here, we present the mathematical expressions used to estimate the robot's state variables in the reorientation experiments, based on the incoming motion capture system (OptiTrack), followed by the impact force estimation. Any rigid body, such as a flying robot, requires twelve variables to define its states in space (\cref*{fig:kinematics}). These are namely three position and three velocity states associated with the translational motion, and three angular position and three angular velocity states associated with the rotational motion.

Assuming the robot is a rigid body, the motion capture system is only capable of tracking the position and attitude angles over time. $\vecsub{P}{\I}=\begin{bmatrix} x & y & z \end{bmatrix} ^\top$ denotes the position of the body-attached frame (subscript $\B$), placed at the robot's COG, with respect to an inertial reference frame (subscript $\I$). $\vec{\Gamma}=\begin{bmatrix} \phi & \theta & \psi \end{bmatrix} ^\top$ defines the body orientation with attitude angles (roll $\phi$, pitch $\theta$, and yaw $\psi$) according to the Euler angle convention \cite{beard2012small}. The linear velocities $\vecsub{V}{\B}=\begin{bmatrix} u & v & w \end{bmatrix} ^\top$ and angular velocities $\vecsub{\Omega}{\B}=\begin{bmatrix} p & q & r \end{bmatrix} ^\top$ of the vehicle, along the body-attached axes, are estimated by
\begin{align}
\vecsub{V}{B}       &= \mat{R}_{\B\I} \dot{\vec{P}}_{I} \ , \\
\vecsub{\Omega}{B}  &= \mat{J} \dot{\vec{\Gamma}} \ ,
\end{align}
where the rotational transformation matrices are
\begin{equation}
\mat{R}_{\B\I} = \begin{bmatrix}
	\cn\theta \cn\psi                   &  \cn\theta \sn\psi                    & -\sn\theta          	\\
	\sn\theta \sn\phi \cn\psi-\sn\psi \cn\phi	&  \sn\theta \sn\phi \sn\psi+\cn\psi \cn\phi	&  \cn\theta \sn\phi		\\
	\sn\theta \cn\phi \cn\psi+\sn\psi \sn\phi	&  \sn\theta \cn\phi \sn\psi-\cn\psi \sn\phi	&  \cn\theta \cn\phi
\end{bmatrix} \ ,
\end{equation}
\begin{equation}
\mat{J} = 
\renewcommand\arraystretch{1.4}
\begin{bmatrix}
	1	&  	   0	             &     	-\sin\theta	   	\\
	0	&  \phantom{-}\cos\phi   & 	\sin\phi\cos\theta	\\
	0	&  -\sin\phi             &  \cos\phi\cos\theta
\end{bmatrix} \ ,
\end{equation}
with $\cn$ and $\sn$ being shorthand notation for $\cos$ and $\sin$, respectively. Multiplying any vector defined in the inertial frame with the $\mat{R}_{\B\I}$ matrix rotates it to the body frame. The derivative of the body position $\dot{\vec{P}}_{I}$ and attitude angles $\dot{\vec{\Gamma}}$ are numerically calculated from the tracked data. 

We estimate the maximum primary impact force, denoted as $F_i$, by analyzing the change in acceleration profile during the primary impact phase (highlighted in red in Figure \ref{fig:reorientation}b). The acceleration in the body frame is calculated using numerical differentiation of the speed profile with the central difference scheme
\begin{equation}
\vecsub{a}{B}^n = \frac{{\vecsub{V}{B}^{n+1} - \vecsub{V}{B}^{n-1}}}{{2\Delta t}},
\end{equation}
where $n+1$ and $n-1$ represent neighboring data points to point $n$, and $\Delta t$ is the time step (equivalent to $\frac{1}{\SI{240}{\hertz}}$). If $m$ denotes the total mass of the vehicle and $a_{i}$ is the magnitude of the maximum acceleration data point, then
\begin{equation}
F_i = m a_i \ . \label{eq:impact force}
\end{equation}

\subsection*{Static perching model}
\label{sec:model}

\bmhead{Model assumptions}
The wing-wrapping model developed is based on some fundamental assumptions, the most significant of which being that the statically perched UAV can be analyzed in a 2D plane rather than in the whole 3D space. The idea is to first analyze all the interactions between the robot and the pole on which it is perched and then move to the 3D space to calculate the maximum weight the system can hold.

Another critical point is that all the different segments are assumed to be touching the pole. This is not always the case as sometimes, one of the segments loses contact with the pole since the configuration with all the segments touching it is not an equilibrium position. Nevertheless, this assumption is valid if no extreme combinations of pole sizes, UAV wingspan, and spring moment are considered.

The torsion springs are modeled to exert loads linearly varying with angle, even if they operate slightly outside manufacturer values for linear operation. The model also assumes a Coulomb friction scheme at the contact points on a macro level and takes no consideration for wet or dry adhesion effects on a micro scale. The total friction force is split unevenly in the horizontal and vertical directions, with its magnitude directly proportional to the applied normal load to the surface and independent of the contact area.


It is also noteworthy that the model only supports symmetrical wing designs, but with a possibility of using uneven segmentation per wing. Even though the robot used in the study has two moving segments of equal length, the number of segments on each wing can be set to any number, and the length of each segment can be selected independently. The model takes the pole and UAV data as inputs, calculates the glider's geometrical configuration, computes all the forces, and determines iteratively whether remaining statically perched on the pole is possible. The data regarding the pole that the models needs are the diameter and the coefficient of static friction only. On the other hand, the data concerning the UAV include all the geometrical and physical parameters as well as the spring information.

\bmhead{Friction model}
Before presenting the step by step iterative force estimation process, it is helpful to make a minor remark. The friction force at each contact point with the pole surface acts in two different directions. One component is tangential ($F_t$) and lies in the $x-y$ plane, and the other is vertical ($F_v$), parallel to the axis of the pole along $z$ axis (see \cref{fig:static model}a). The first one keeps the UAV in place, avoiding slipping and detachment from the pole, while the latter is the force that is actively overcoming gravity, preventing the robot from falling.

Using the Coulomb friction model
\begin{equation}
    F_f = \mu_s F_n \ , \label{eq:coulomb_friction}
\end{equation}
where $F_f$ is the total friction force split into two components. Using two figurative friction coefficients and leveraging the vectorial sum:
\begin{equation}\label{eq:friction_comps}
    \begin{cases}
    \vecsub{F}{f} = \vecsub{F}{t} + \vecsub{F}{v} \ , \\
    F_t = \mu_t F_n \ , \\
    F_v = \mu_v F_n \ .
    \end{cases}
\end{equation}

Given that the two components will always be perpendicular to each other, the relationship between the two friction coefficients should satisfy
\begin{equation}\label{eq:friction_coeffs}
    \mu_s^2 = \mu_t^2+\mu_v^2 \ .
\end{equation}

\bmhead{Solving scheme}
We first solve a purely geometric problem to find the contact points. The model positions a pole of diameter $\diameter$ with its center at the origin of the $x-y$ plane (\cref{fig:static model}a). Then, it assumes that the UAV and the pole will be tangent at each point of contact. The fuselage is always positioned tangent to the lowest point in the circumference. Then all the segments are added, starting from the innermost ones closest to the fuselage, leveraging the tangent constraint and the fact that they share a hinge with the fuselage or the previous segment. Once the relative positions of the pole and the segments are determined, we know the angles at each hinge and can compute the moments provided by the torsion springs using
\begin{equation}
    M_s = k_s \theta_h \ ,
\end{equation}
where $k_s$ and $\theta_h$ are the stiffness of the torsion spring and hinge angle, respectively. The model solves the planar multibody problem starting from the outermost segment (denoted by number $2$ on \cref{fig:static model}) since it is the only one affected by a single spring. Therefore, looking at this segment only, the normal force $F_{n_2}$ exerted from the pole to the UAV is calculated using the rotational equilibrium with respect to the hinge. Once the normal force is known, the in-plane component of the friction force $F_{t_2}$ is computed using the corresponding tangential friction coefficient $\mu_t$ as such
\begin{align}
    F_{n_2} &= \dfrac{M_{s_2}}{l_{2,2}} \ , \\
    F_{t_2} &= \mu_t F_{n_2} \ ,
\end{align}
where $l_{2,2}$ is the corresponding length of segment $2$ from the hinge location $h_2$ to its contact point $c_2$. The model then advances to the next segment, solved similarly, always leveraging the moment equilibrium. The considered subsystem can be solved at each instance since the forces acting on the previous segments are already calculated earlier. For the sake of brevity, we only present the the moment equilibrium equation in vector form, from which $F_{n_1}$ and $F_{t_1}$ can be computed
\begin{equation}
    \vecsub{M}{s_1} + 
    \vecsub{l}{1,1} \times \vecsub{F}{n_1} +
    \vecsub{l}{1,2} \times \vecsub{F}{n_2} +
    \vecsub{l}{1,2} \times \vecsub{F}{t_2} = 0 \ ,
\end{equation}
\begin{equation}
    F_{t_1} = \mu_t F_{n_1} \ .
\end{equation}

All the normal and in-plane friction forces concerning a single wing are determined. Then, leveraging the symmetrical constraint mentioned, the other half of the system is solved, followed by the final force and moment equilibrium on the fuselage to determine the last set of normal and tangential forces.

\bmhead{Algorithmic process and friction split}
The only critical element remained in the modeling process is knowing how the friction splits between the planar and vertical components. This cannot be estimated a priori knowing only the data regarding the robot and the pole. To solve this issue, the model sweeps through the whole range of possible splits between the horizontal and vertical friction coefficients and finds which combinations bring to a possible equilibrium position. It then selects the solution as the case that minimizes the net friction coefficient.

The flowchart in \cref*{fig:flowchart} is presented to illustrate this part of the model more clearly. The whole process is initialized by calculating the weight of the UAV ($W$) and setting the fraction of friction acting in the $x-y$ plane ($\mu_{h,\%}$) to zero percent. Then, this value will be increased by a percentage at each iteration until it reaches \SI{100}{\percent}. Inside each iteration, another process is triggered. This second one is initiated by setting to zero the total friction coefficient percentage ($\mu_\%$) with respect to the maximum possible one (the static friction coefficient, $\mu_s$). Therefore, the total friction coefficient ($\mu$) is null and the vertical force, given by the vertical component of the friction force ($F_v$), is also null. Then, a new cycle starts. First, the two friction coefficients ($\mu_h$ and $\mu_v$) are calculated. Then, the model solves the system by leveraging the rotational equilibrium and determines whether the specific friction split considered is a possible solution or not. If a solution is found, the cycle stops, while, if this is not the case, the total friction coefficient percentage ($\mu_\%$) is increased, and a new configuration is studied. This process is repeated until a solution is found or the total friction coefficient reaches the maximum possible one ($\mu_s$). At the end of these nested loops, the model scans all the found possible solutions and selects the friction split that minimizes the total friction coefficient needed ($\mu$) to remain perched.

\subsection*{Wing segmentation sizing}
\label{sec:segmentation}

The static model can aid with proper dimensioning of the wings at the design stage. While in theory, it is the wingspan that sets the range of poles on which the robot can potentially perch, segmentation of the wing affects the static load carrying capacity. The model serves as a valuable tool during the design stage in selecting the right segment size for a given number of segments. For PercHug, we selected two equal folding segments per wing and explored three configurations by varying the fuselage and moving segment widths to achieve a constrained wingspan of \SI{960}{\milli\meter}. 

These considered wing configurations correspond to folding segments of width \SIlist{205;195;185}{\milli\meter}. Based on simulation results, we selected the second configuration which outperformed the others, exhibiting higher static payload capacity on most poles. This finding is illustrated in \cref*{fig:configurations}-c, where it is evident that the third configuration had limited success on poles, while the first and second configurations performed well, with the second one being the preferred choice. The simulation results are supported by experimental verification with the selected wing configuration (see \nameref{sec:static perching experiments} for more detail).

\subsection*{Static perching experiments}
\label{sec:static perching experiments}

We validated the predicted wing design (see \nameref{sec:segmentation} for configuration selection) with experiments of a physical prototype weighing \SI{325}{\gram} attached to poles of different sizes and materials. The testing strategy involved placing the prototype robot, equipped with hook-less perching wings, on various poles and trees within the range specified by the model. We then gradually increased the weight in steps of \SI{100}{\gram} until the point of falling, allowing us to estimate the maximum static perching payload. The surface texture and specifications of the poles are listed in \cref{fig:static perching}a and b (see \nameref{sec:friction measurement} for the methods used to estimate coefficient of static friction). The tested objects included two indoor poles with diameters \SI{250}{\milli\meter} and \SI{315}{\milli\meter}, covered in copy paper, rubber pad, and paper towel (designated as poles I, II, and VI-IX), a \SI{260}{\milli\meter} bamboo tree guard (IV), smooth and rough concrete columns of \SI{350}{\milli\meter} (III and V), and six trees (X-XV) with diameters ranging from \SIrange{265}{360}{\milli\meter}.

\subsection*{Friction coefficient measurement}
\label{sec:friction measurement}

One of the two essential pieces of information needed to run the static perching model is the value of the static friction coefficient between EPP and the surface of the pole. The way to measure this value changes based on the specific case considered. For poles I, II, and VI-IX, the friction coefficient measurement was straightforward since the materials used to cover the indoor poles could be detached from them and used for measuring purposes. We used two methods to estimate the static friction coefficients on these poles. First method included placing the material of interest on a flat surface and positioning a block of EPP foam (with known weight, $m$) on top. We then used a force measuring device, a spring balance, to gradually increase the pulling force on the foam block ($F_{pull}$) until it started slipping. The value of the static friction coefficient was then given by
\begin{equation}\label{eq:static_coeff}
    \mu_s = \frac{F_{pull}}{m g} \ .
\end{equation}

The second method employed was based on the estimation of friction angle $\theta$, i.e., the maximum possible angle that still prevents sliding. The first material is placed at some variable angle with respect to the horizontal direction, the foam block is positioned on top, and the angle is increased until the block starts sliding. By measuring the angle, the static friction coefficient can be estimated by
\begin{equation}\label{eq:friction_angle}
    \mu_s = \tan\theta \ .
\end{equation}

We used both techniques for all the cases with the possibility of removing the surface material from the pole. We measured the static friction coefficient 10 times with each method. The final value was computed as the average of all measurements. 

For the rest of the poles (mainly trees) in which the surface material could not be detached from the pole itself, the measurements had to be taken on a vertical surface. Since no standard testing procedure was found in the literature, we proposed a new technique to do so. The concept involves pushing a piece of the material of interest with known weight ($mg$) against the vertical surface of the pole with a known force and pulling it upward ($F_{pull}$) using a force measuring device until sliding. An essential requirement for such a technique is the ability to create a known normal force using elements that do not affect the object under measurement with their masses or other forces that could change the balance between them. 

We designed the measuring tool in \cref*{fig:friction tool} to tackle such an issue. It is composed of an aluminum frame with adjustable arms that tightly grips the pole and a foam block that is pressed against the surface of the pole using a set of four linear springs. The linear springs are placed inside small channels to keep them straight along the horizontal direction. Since the springs are the only connection between the block and the aluminum structure, the weight of the support structure is not passed to the block. Therefore, the measurement of the friction is not affected. The compression of the springs can be calculated by measuring the distance between the foam block and the fixed plastic plate, relative to their original lengths. Consequently, knowing the elastic pressing force, the static friction coefficient is computed by
\begin{equation}\label{eq:friction_coeff_vertical}
    \mu_s = \frac{F_{pull}-mg}{k\Delta l}.
\end{equation}
where $k$ is the spring stiffness, and $\Delta l$ is the compression amount.


\subsection*{Dynamic perching experiments}
\label{sec:dynamic perching experiments}

With the trigger strategy set to release upon primary impact, we used the PercHug prototype to conduct a minimum of five perching experiments on each tree for the two selected nose configurations, i.e., the standard upturned nose and the elastic extension with $D=\SI{0.233}{\newton\meter\squared}$. This resulted in a total of 80 perching trials, approximately 40 tests for each nose type across the six trees. The perching events were recorded in slow-motion at \SI{240}{\hertz}, and the speed and pitch angle data were extracted using Physlets Tracker software with the assistance of colored markers placed on PercHug.

For the perching characterization, we only considered experiments that met specific criteria. One crucial requirement was the successful execution of a reorientation maneuver, as unsuccessful reorientation would inevitably lead to unsuccessful perching. Previous findings of the \restructure{\nameref{sec:reorientation}} section indicated that a minimum impact angle of \SI{15}{\degree} and \SI{8}{\degree} was necessary to ensure successful reorientation with standard and elastic noses, respectively (\cref{fig:reorientation}c). Therefore, trials that did not meet these conditions were excluded. Additionally, tests in which the impact occurred on the fuselage or wings instead of the nose, resulting in no or unsuccessful reorientation, were also excluded. As a result, 16 out of 80 trials were disregarded.

\backmatter

\section*{Declarations}

\bmhead{Acknowledgments}
This work was supported in part by NCCR Robotics, a National Centre of Competence in Research, funded by the Swiss National Science Foundation (grant no: 51NF40\_185543), and by the European Union's Horizon 2020 research and innovation program through the AERIAL-CORE project (grant no: 871479).

\bmhead{Supplementary information}
The article contains supplementary material (a list of files available in the accompanying PDF file).

\bmhead{Author contributions}
M.A., W.S., A.J.I, and D.F. conceptualized the ideas and organized the study. M.A. and M.B. designed and fabricated the robot prototypes. M.A. conducted the reorientation experiments and analyzed the data (with input from H.V.P. and W.S.). M.B. developed the static model and performed static perching experiments (with input from M.A. and W.S.). M.A. and M.B. conducted the dynamic perching experiments with the PercHug robot and analyzed the data. M.A., M.B., H.V.P., W.S., A.J.I, and D.F. wrote and revised the manuscripts.

\bmhead{Data availability}
All data needed to evaluate the conclusions of the paper are available in the main manuscript and the supplementary information.

\bmhead{Competing interests}
The authors declare no competing interests.


\bibliography{References} 


\end{document}